# Learning Implicit Generative Models with the Method of Learned Moments

**Suman Ravuri** [1]  **Shakir Mohamed** [1]  **Mihaela Rosca** [1]  **Oriol Vinyals** [1]

## Abstract

We propose a method of moments (MoM) algorithm for training large-scale implicit generative models. Moment estimation in this setting encounters two problems: it is often difficult to define the millions of moments needed to learn the model parameters, and it is hard to determine which properties are useful when specifying moments. To address the first issue, we introduce a *moment network*, and define the moments as the network's hidden units and the gradient of the network's output with respect to its parameters. To tackle the second problem, we use asymptotic theory to highlight desiderata for moments – namely they should minimize the asymptotic variance of estimated model parameters – and introduce an objective to *learn* better moments. The sequence of objectives created by this *Method of Learned Moments* (MoLM) can train high-quality neural image samplers. On CIFAR-10, we demonstrate that MoLM-trained generators achieve significantly higher Inception Scores and lower Fréchet Inception Distances than those trained with gradient penalty-regularized and spectrally-normalized adversarial objectives. These generators also achieve nearly perfect Multi-Scale Structural Similarity Scores on CelebA, and can create high-quality samples of $128 \times 128$ images.

## 1. Introduction

The method of moments (MoM) is an ancient principle of learning (Pearson, 1893; 1936). At its heart lies a simple procedure: given a model with parameters $\theta$, estimate $\theta$ such that the moments — or more generally feature averages — of the model match those of the data. While the technique is simple and yields consistent estimators under weak conditions, other properties of the moment estimator are less desirable. Moment estimators are often biased, sometimes lie outside the parameter space (such as negative probabilities), and, unless the model is in an exponential family, are less statistically efficient than maximum likelihood estimators. It is perhaps this last property that has relegated the method of moments to a niche technique.

There are, however, situations in which moment estimation is preferable to maximum likelihood estimation (MLE). One is when MLE is more computationally challenging than MoM. For example, one can scale training of Latent Dirichlet Allocation to large datasets by moment estimation using the first three order moments (Anandkumar et al., 2012a). Second, for latent variable models more generally, maximum likelihood estimation using the EM algorithm results in a local optimum, while MoM enjoys stronger formal guarantees (Anandkumar et al., 2012b). Finally, one can use moment estimation to determine model parameters in settings where likelihoods are unnatural. Instrumental variable estimation, an example of MoM, is used to learn parameters in supply and demand models (Wright, 1928).

We study another scenario: when data come from unknown or difficult-to-capture likelihood models. Data such as images, speech, or music often arise from complicated distributions, and for image data in particular, models based on likelihoods often yield low-quality samples. Researchers interested in generating more realistic samples have shifted their efforts to training neural network samplers with alternative losses. They have studied training these implicit generative models with the Wasserstein distance (Arjovsky et al., 2017), Maximum Mean Discrepancy (MMD) (Li et al., 2015; Dziugaite et al., 2015; Sutherland et al., 2016), and other divergences (Nowozin et al., 2016; Mao et al., 2016). While direct minimization of these distances or divergences – namely Wasserstein (Salimans et al., 2018), and MMD (Li et al., 2015; Dziugaite et al., 2015), Cramér (Bellemare et al., 2017) – in pixel space has led to poor sample quality, indirect minimization using adversarial training (Goodfellow et al., 2014) dramatically improves samples.

We pursue an alternative strategy: we explicitly define our moments and train them so that the moment estimators are statistically efficient. This choice creates two practical problems. The first is that traditional neural samplers, such as Deep Convolutional Generative Adversarial Networks (DC-GANs) (Radford et al., 2015), have millions of parameters, and typically one needs at least one moment per parameter to train the model. Second, it is not obvious how to choose or train moments so that they are statistically efficient.

[1]DeepMind London, N1C 4AG, UK . Correspondence to: Suman Ravuri <suman.ravuri@google.com>.

To address the first issue, we introduce a *moment network*, whose activations and gradients constitute the set of moments with which we train the generator. To tackle the second problem, we appeal to asymptotic theory to determine desiderata for moments (namely that they minimize the asymptotic variance of estimated model parameters) and explicitly specify and learn moments to train the generator. It is this theory that is used in the literature of the generalized method of moments in econometrics to reweight moments to make estimators more statistically efficient (Hansen, 1982; Hall, 2005).

We make the following contributions:

- We demonstrate that method of moments can scale to neural network models with tens of millions of parameters.
- We highlight the importance of statistical efficiency in method of moments and provide a method for learning moments such that moment estimators minimize asymptotic variance.
- We show that implicit generative models trained with our algorithm, the *Method of Learned Moments*, generate samples that are as good as, or better than, models that use adversarial learning, as measured by standard metrics.

## 2. The Method of Learned Moments

### 2.1. A Review of the Method of Moments

Suppose our data are drawn i.i.d. from $x_i \sim p^*$, and our samples $s$ are drawn i.i.d. from a model $p_\theta$, whose parameters $\theta \in \mathbb{R}^n$ we wish to learn. Method of Moments (MoM) estimation requires us to define feature functions $\Phi(x) \in \mathbb{R}^k$ and an associated moment function $m(\theta) := \mathbb{E}_{p_\theta(s)}[\Phi(s)]$. The moment estimator $\hat{\theta}_N$ matches the moment function with the feature average over the data:

$$m(\theta) = \frac{1}{N} \sum_{i=1}^{N} \Phi(x_i) \quad \theta = \hat{\theta}_N$$

If $p_\theta \xrightarrow{d} p^*$ for some $\theta^*$, the moments exist, and $m(\theta) \neq m(\theta^*) \;\; \forall \theta \neq \theta^*$, this feature matching will yield a consistent estimator of $\theta^*$. When we have access to the likelihood, we can recover the maximum likelihood estimate by setting $\Phi(x) = \nabla_\theta \log p_\theta(x)$ and noting that the expected value of the score function is zero at $\theta^*$. Since maximum likelihood estimators are asymptotically efficient, they are generally preferable to their moment counterparts (Van der Vaart, 1998, Ch. 8).

With implicit generative models (Mohamed and Lakshminarayanan, 2016), we no longer have explicit access to the likelihood. This precludes straightforward application of maximum likelihood. Instead, we have indirect access to $p_\theta$ through a parametric sampler $g_\theta(z) \sim p_\theta$. MoM estimation, however, is still applicable by replacing the moment function $m(\theta) := \mathbb{E}_{p_\theta(s)}[\Phi(s)]$ with $m(\theta) := \mathbb{E}_{p(z)}[\Phi(g_\theta(z))]$, where $z$ is a draw from a prior distribution $p(z)$, such as a Gaussian or uniform distribution. If the generative model is sufficiently expressive to model the data distribution, the same regularity conditions on $\Phi$ will ensure a consistent estimator of generator parameters $\theta$.

While consistency is guaranteed, the asymptotic efficiency argument of maximum likelihood implies that the specification of features $\Phi(x)$ affects the quality of the learned model parameters. One desirable aspect of the moment-matching framework is a developed asymptotic theory that provides large-sample behavior of the moment estimator. Intuitively, it tells us that our estimated generator parameters after $N$ datapoints is roughly distributed as a Gaussian with mean $\theta^*$ and variance $V/N$. Minimizing $V$ makes estimation of generator parameters more data-efficient and depends on the quality of $\Phi(x)$. The following theorem allows us to connect our choice of moments with its asymptotic variance.

**Theorem 1** (Asymptotic Normality of Invertible Moment Functions). *Let* $m(\theta) = \mathbb{E}_{p(z)}[\Phi(g_\theta(z))]$ *be a one-to-one function on an open set* $\Theta \subset \mathbb{R}^d$ *and continuously differentiable at* $\theta^*$ *with nonsingular derivative* $G = \nabla_\theta \mathbb{E}_{p(z)}[\Phi(g_{\theta^*}(z))]$. *Then assuming* $\mathbb{E}_{p(z)}[\|\Phi(g_{\theta^*}(z))\|^2] < \infty$, *moment estimators* $\hat{\theta}_N$ *exist with probability tending to one and satisfy*

$$\sqrt{N}(\hat{\theta}_N - \theta^*) \to \mathcal{N}(0, G^{-1} \Sigma G^{-\mathsf{T}})$$
$$\Sigma := cov(\Phi(g_{\theta^*}(z)))$$

*Proof.* See Theorem 4.1 in Van der Vaart (1998) ☐

Minimizing this asymptotic covariance requires balancing $G$ and $\Sigma$. $G$ asserts that one should maximize the difference in features between the optimal generator parameters and those a small distance away, while $\Sigma$ expresses that one should minimize the covariance of the features for the optimal generator parameters.

While this theorem requires the restrictive condition of invertible moment functions, the theorem in this ideal setting allows us to design statistically efficient moments. Moreover, in section 2.2.1 we later relax the assumptions of invertibility while showing the design choices still hold.

To obtain better moments, we will explicitly create parametric moments and optimize those moments to be more statistically efficient. This approach introduces two hurdles: 1) defining millions of sufficiently different moments and 2) creating an objective to learn desirable moments. Our practical contribution comprises how we solve these two issues.

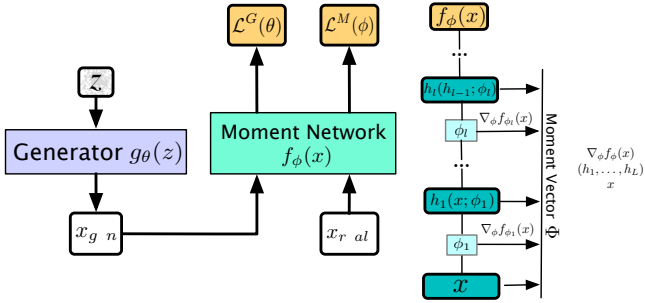

*Figure 1.* Illustration of method of learned moments architecture.

## 2.2. From Theory to Practice

### 2.2.1. MOMENT SPECIFICATION

Moment specification is an exercise in ensuring consistency of the moment estimator. The assumption in Theorem 1 that the moment function is invertible is rather restrictive. One can instead weaken it to require identifiability: $m(\theta) = \mathbb{E}_{p^*(x)}[\Phi(x)]$ iff $\theta = \theta^*$. Even this condition, however, is difficult to verify in practice. We thus resort to a local identifiability assumption used in econometrics: that there are more moments than model parameters, and that $G$ is full rank (Hall, 2005, Ch. 3). For linear models, this ensures global identifiability, and for nonlinear ones, this heuristic tends to work well in the literature.

Explicitly specifying moments for neural samplers seems especially daunting, however, since generally one needs at least as many moments as sampler parameters for moment matching to yield a consistent estimator of the model parameters.[1] For the DCGAN architecture with batch normalization, depending on the resolution of the dataset, the number of feature maps, and kernel size, one would need up to 20 million moments.

Our solution is to create a *moment network* $f_\phi(x)$ and define moments as:

$$\Phi(x) = [\nabla_\phi f_\phi(x), x, h_1(x) \ldots h_{L-1}(x)]^\mathsf{T}$$

where $h_i(x)$ are the activations for layer $i$. $h_L(x)$ is implicitly included in the gradient. As long as the moment network has as many parameters as the generator, there are enough moments to train the model. Although we cannot ensure that $G$ is full rank, we typically scale the moment network to produce between 1.5 and 5 times as many moments as generator parameters. We use both the gradients and activations since they encode different inductive biases. We discuss those biases in Section 2.5.

[1] Moreover, neural samplers also suffer from identifiability issues. For example, one can express the same function by permuting nodes and associated parameters. We assume that, given an initialization $\theta_0$, only one $\theta^*$ is achievable using gradient descent, but we leave a more rigorous argument for future work.

Since the moment function is not invertible, we replace the moment-matching objective with the squared error loss between moments of the data and samples:

$$\mathcal{L}^G(\theta) = \frac{1}{2} \left\| \frac{1}{N} \sum_{i=1}^{N} \Phi(x_i) - \mathbb{E}_{p(z)}[\Phi(g_\theta(z))] \right\|_2^2 \quad (1)$$

Figure 1 illustrates our setup. In practice, we take Monte Carlo estimates of the sampler expectation. The change in objective function and Monte Carlo estimate induce a change in the asymptotics of this modified moment estimator. We address the effect of these changes in Section 2.3.

### 2.2.2. LEARNING EFFICIENT MOMENTS

It may seem plausible that given enough moments parameterized by random $\phi$, the underlying objective is sufficiently good in practice to train neural samplers. Unfortunately, as we show in section 4, sample quality is poor. Theorem 1 allows us to diagnose the problem. $G$, the Jacobian of the moment estimator with respect to the optimal generator parameters is not sufficiently "large". From the definition of the Jacobian:

$$G(\theta - \theta^*) = \mathbb{E}_{p(z)}[\Phi(g_\theta(z))] - \mathbb{E}_{p(z)}[\Phi(g_{\theta^*}(z))] + o(\|\theta - \theta^*\|)$$

For $\theta$ near the optimum, maximizing the difference in expected features also maximizes the "directional Jacobian" $G$. One does not expect, however, that moments produced for random $\phi$ to maximize this difference. This motivates learning $\phi$.

Of course, since we do not have access to $\theta^*$, exact computation of the asymptotic variance components $G$ and $\Sigma$ is impossible. Under the assumption that the generator is sufficiently expressive to represent the data distribution, then $p(x) \overset{d}{=} p(g_{\theta^*}(z))$, $\mathbb{E}_{p(z)}[\Phi(g_{\theta^*}(z))] = \mathbb{E}_{p(x)}[\Phi(x)]$ and $\Sigma = cov[\Phi(x)]$. We make a first approximation of maximizing this directional Jacobian:

$$G(\theta - \theta^*) \approx \mathbb{E}_{p(z)}[\Phi(g_\theta(z))] - \mathbb{E}_{p(x)}[\Phi(x)]$$

In early experiments optimizing this difference, moments became correlated and as a result the estimator of $\theta$ was no longer consistent. Thus, we make a second approximation. Inspired by the work of Jaakkola and Haussler (1999) in extracting feature vectors from auxiliary models for use in a linear SVM classifier, Tsuda et al. (2002) proposed the gradient of the log-odds ratio of a probabilistic binary classifier as a model-dependent feature. The authors empirically and theoretically show improved binary classification (and thus separability).

Applying this idea to our method, we perform binary classification on real images and our samples. Denoting

$D_\phi(x) = P(y = 1|x)$ as the probability that $x$ is a real image, then the tangent of posterior odds is:

$$\nabla_\phi \log \frac{P(y=1|x)}{P(y=-1|x)} = \nabla_\phi \log \frac{\sigma(f_\phi(x))}{1 - \sigma(f_\phi(x))} = \nabla_\phi f_\phi(x)$$

where $\sigma$ is the logistic sigmoid function. When the log-odds ratio is included in the feature, the resulting kernel $K(x, y) = \Phi(x)^\mathsf{T} \Phi(y)$ is known as Tangent of Posterior Odds Kernel (Tsuda et al., 2002).

To help control $\Sigma$, we add a quadratic penalty on the squared norm of the minibatch gradient, so that the average squared moment is close to 1. The moment objective is now:

$$\mathcal{L}^M(\phi) = \mathbb{E}_{p(x)}[\log D_\phi(x)] + \mathbb{E}_{p(z)}[\log(1 - D_\phi(g_\theta(z)))]$$
$$+ \lambda \left( \frac{\|\mathbb{E}_{p(x)}[\nabla_\phi f_\phi(x)]\|^2}{k} - 1 \right)^2, \quad \nabla_\phi f \in \mathbb{R}^k$$

We do not include regularization on the hidden units as they represent a small percentage of our moments and its regularization yielded no difference in performance. A more correct penalty term is $\left(\frac{1}{k}\mathbb{E}_{p(x)}[\|\nabla_\phi f_\phi(x)\|^2] - 1\right)^2$ (to make sure the second moments are close to 1), but we found no performance benefit from the extra computational cost.

The upshot of learning moments is that each parameterization of $f_\phi$, subject to regularity conditions, produces a consistent estimator of $\theta$. On the other hand, each set of moments is not necessarily asymptotically efficient, as they rely on poor estimates of $G$. So, we learn better $\Phi$ iteratively (usually every 1,000-2,000 generator steps), and then match those moments. Algorithm 2.5 describes the proposed method.

### 2.3. Refinement of Asymptotic Theory

Note that while we defined the asymptotics for the feature-matching objective and used that to learn moments, our loss is actually the squared error objective in Equation 1. Although the same tradeoff applies to that objective, its asymptotic variance is somewhat more complicated. To develop the asymptotics of that expression, for clarity let us define the moment function:

$$m_N(x_{1,...,N}, \Phi, \theta) = \frac{1}{N} \sum_{i=1}^{N} \Phi(x_i) - \mathbb{E}_{p(z)}[\Phi(g_\theta(z))]$$

The asymptotics of the weighted squared loss function:

$$\mathcal{L}^G(\theta) = m_N(x_{1,...,N}, \Phi, \theta)^\mathsf{T} W m_N(x_{1,...,N}, \Phi, \theta)$$

are:

**Theorem 2** (Asymptotic Normality of Squared Error Functions). *Under the consistency and asymptotic normality conditions in Appendix B.1, the estimator satisfies:*

$$\sqrt{N}(\hat{\theta}_N - \theta^*) \to \mathcal{N}(0, V_{SE})$$
$$V_{SE} := (G^\mathsf{T} W G)^{-1} G^\mathsf{T} W \Sigma W G (G^\mathsf{T} W G)^{-1}$$

*where* $\Sigma := cov(\Phi(g_{\theta^*}(z)))$.

*Proof.* Theorem 3.2 of Hall (2005) ☐

When $W \propto \Sigma^{-1}$, then

$$\sqrt{N}(\hat{\theta} - \theta^*) \to \mathcal{N}(0, (G^\mathsf{T} \Sigma^{-1} G)^{-1})$$

It can be shown that this is the optimal weighting matrix. The inverse of this matrix is known as the Godambe Information Matrix (Godambe, 1960) and serves as a generalization of the Fisher Information Matrix.

Of course, the above theorem presupposes that we can analytically calculate $\mathbb{E}_{p(z)}[\Phi(g_\theta(z))]$. In implicit generative modeling, however, we only have access to $\frac{1}{K} \sum_{k=1}^{K} \Phi(g_\theta(z_k))$. It turns out we only pay a constant factor penalty for sampling. More specifically, suppose our moment function is now:

$$\hat{m}_N(x_{1,...,N}, \Phi, \theta) = \frac{1}{N} \sum_{i=1}^{N} \Phi(x_i) - \frac{1}{\mathcal{T}(N)} \sum_{k=1}^{\mathcal{T}(N)} \Phi(g_\theta(z_{i,k}))$$

where $\mathcal{T}(N)$ is the number of samples used to estimate generator moments for $N$ points in the dataset. Then the asymptotic variance of this method, known as the simulated method of moments (Hall, 2005, Ch. 10), is:

**Theorem 3** (Asymptotic Normality of Simulated Method of Moments). *Suppose that* $\frac{\mathcal{T}(N)}{N} \to K$ *as* $N \to \infty$. *Assuming the conditions in Appendix B.2, then* $\hat{\theta}_N$ *satisfies.*

$$\sqrt{N}(\hat{\theta}_N - \theta^*) \to \mathcal{N}\left(0, \left(1 + \frac{1}{K}\right) V_{SE}\right)$$

*Proof.* See Duffie and Singleton (1993) ☐

### 2.4. Computational Considerations

The gradient for the squared-error objective is:

$$\nabla_\theta \mathcal{L}^G(\theta) = \frac{1}{K} \sum_i J_i^\mathsf{T} \bar{m}(x, \Phi, \theta)$$

where $J_i := \nabla_\theta \Phi(g_\theta(z_i))$ is a Jacobian matrix and $\bar{m}(x, \Phi, \theta) = \frac{1}{N} \sum_{n=1}^{N} \Phi(x_i) - \frac{1}{K} \sum_{i=1}^{K} \Phi(g_\theta(z_i))$.

When using only gradient features, one can speed up gradient computation by $\sim 20\%$ by using a Hessian-vector product-like trick (Pearlmutter, 1994; Schraudolph, 2002).

Note that Hessian-vector products are defined as:

$$Hv = \left[ \begin{array}{c|c} H_{\theta\theta} & J^\mathsf{T} \\ \hline J & H_{\phi\phi} \end{array} \right] [v]$$

---

**Algorithm 1** Method of Learned Moments

**Input:** Learning rate $\alpha$; number of objectives $N_o$; number of moment training steps $N_m$; number of generator training steps $N_g$; norm penalty parameter $\lambda$

Initialize generator and moment network parameters to $\theta$ and $\phi$ respectively

**for** $n = 0, \ldots, N_o$ **do**
  **for** $n = 1, \ldots, N_m$ **do**
    $d_\phi \leftarrow \nabla_\phi \mathcal{L}^M(\phi)$
    $\phi_{t+1} \leftarrow \phi_t - \alpha \cdot \text{AdamOptimizer}(\phi_t, d_\phi)$
  Calculate $\frac{1}{N} \sum_i \Phi(x_i)$ over the entire dataset.
  **for** $n = 1, \ldots, N_g$ **do**
    $d_\theta \leftarrow \nabla_\theta \mathcal{L}^G(\theta)$
    $\theta_{t+1} \leftarrow \theta_t - \alpha \cdot \text{AdamOptimizer}(\theta_t, d_\theta)$

---

Let $v$ be defined as a partitioned vector as follows:

$$v = \begin{bmatrix} 0 \\ \overline{m}(x, \Phi, \theta) \end{bmatrix} \text{ Then: } Hv = \begin{bmatrix} J^\mathsf{T} \overline{m}(x, \Phi, \theta) \\ H_{\phi\phi} \overline{m}(x, \Phi, \theta) \end{bmatrix}$$

Performing a Hessian-vector computation only through the moment network provides the desired gradients.

### 2.5. Moment Architectures and Inductive Biases

When choosing a particular moment architecture, we implicitly specify the inductive biases of our features. Researchers typically design such architectures such that the forward pass encodes properties such as translational invariance and local receptive fields. These are properties of the hidden units. Gradients, however, often encode far different properties. For the convolutional moment architectures used in our experiments, gradient features encode global properties of the data. To see this, note that the output of a convolution is $y_{l,m,n} = \sum_a \sum_b \sum_c \phi_{a,b,c,n} x_{l+a,m+b,c}$. Its partial derivatives are $\frac{\partial y_{l,m,n}}{\partial \phi_{a,b,c,n}} = x_{l+a,m+b,c}$. Applying chain rule gives us the partial derivative with respect to $o = f_\phi(x)$

$$\frac{\partial f_\phi(x)}{\partial \phi_{a,b,c,n}} = \sum_l \sum_m \frac{\partial f_\phi(x)}{\partial y_{l,m,n}} x_{l+a,m+b,c}$$

Tying weights, which helps us encode local properties of the image in the forward pass, instead gives us global properties in the backward pass. Hence, we augment gradient features with hidden units to balance both local and global structure.

## 3. Connection to Other Methods

### 3.1. Maximum Mean Discrepancy

The method of moments probably bears the closest relationship to Maximum Mean Discrepancy (MMD)[2]. One can consider method of moments as embedding a probability distribution into a finite-dimensional vector. MMD, on the other hand, embeds a distribution into an infinite-dimensional vector. By enforcing $\phi \in \mathbb{L}^2$, one can calculate this "infinite-moment" matching loss as sum of expectations of kernels:

$$\text{MMD}^2(\theta) = \sum_{i=1}^\infty (\mathbb{E}_{p(x)}[\phi_i(x)] - \mathbb{E}_{p(z)}[\phi_i(g_\theta(z))])^2$$

$$= \mathbb{E}[K(x, x')] - 2\mathbb{E}[K(x, g_\theta(z))] + \mathbb{E}[K(g_\theta(z), g_\theta(z'))]$$

---

[2] We assume the Reproducing Kernel Hilbert Space (RKHS) version of MMD; i.e., the function class is $\mathcal{F} = \{f \| \|f\|_{\mathcal{H}} \leq 1\}$.

Furthermore, if the kernels are characteristic, then $\text{MMD}^2$ defines a squared distance (Gretton et al., 2012). Method of moments, on the other hand, is only able to distinguish between probability distributions specified by the model.

Despite robust theory, sample quality has lagged behind adversarial methods, especially if radial basis function (RBF) kernels are used. An explanation perhaps lies in the analysis of $\text{MMD}^2$ loss as spectral-domain moment matching.

**Proposition 1.** *Suppose the kernel function $K(x, y) = K(x - y)$ is real, shift-invariant, Bochner integrable, and without loss of generality $K(0)=1$. Then:*

$$\mathbb{E}_{p(x,x')}[K(x, x')] - 2\mathbb{E}_{p(x,y)}[K(x, y)] + \mathbb{E}_{p(y,y')}[K(y, y')]$$

$$= \mathbb{E}_{p(w)}[(\mathbb{E}_{p(x)}[\cos(\omega^\mathsf{T} x)] - \mathbb{E}_{p(y)}[\cos(\omega^\mathsf{T} y)])^2]$$

$$+ \mathbb{E}_{p(w)}[(\mathbb{E}_{p(x)}[\sin(\omega^\mathsf{T} x)] - \mathbb{E}_{p(y)}[\sin(\omega^\mathsf{T} y)])^2] \quad (2)$$

*where $p(\omega)$ is a probability measure specified by the kernel.*

*Proof.* See Appendix C.1 □

Crucially, for radial basis function kernels, $p(\omega) \propto \exp(-\frac{1}{2}\sigma^2 \|\omega\|^2)$. For high-dimensional data, unless the data lie on a spherical shell of appropriate radius, then one likely needs many samples to accurately approximate $\text{MMD}^2$ distance.

It may be the poor spectral properties of the RBF kernels that have led to poorer samples. More recent work has focused on other kernels – such as sums of RBF kernels at different bandwidths (Sutherland et al., 2016) and rational quadratic kernels (Bińkowski et al., 2018) – and indeed using those kernels improved sample quality. In fact, the proposed Coulomb GAN (Unterthiner et al., 2017) shares a deeper relationship with MMD. It directly minimizes MMD loss using a version of the rational quadratic kernel known as the Plummer kernel to estimate $f$, and further introduces a discriminator to model the scaled witness function $f^*$. The upshot is that the generator loss approximates a high-sample biased estimate of MMD loss; see Appendix C.2 for details.

### 3.2. Adversarial Training

Recent work (Liu et al., 2017) has shown that in practical settings, many GAN objectives are better expressed as gen-eralized moment matching as discriminators have limited capacity. Given this viewpoint, can asymptotic theory tell us anything about the statistical efficiency of adversarial networks?

Unfortunately, since there are an "infinite" number of moments, we cannot directly apply the above asymptotic theory. We do note that the inner maximization step indirectly "maximizes" $G$. For clarity, consider the inner maximization step of the Wasserstein GAN:

$$\mathcal{L}(\phi) = \min_\theta \max_\phi \mathbb{E}_{p(x)}[f_\phi(x)] - \mathbb{E}_{p(z)}[f_\phi(g_\theta(z))]$$

Under the assumption that the generator at $\theta^*$ is sufficiently expressive to model the data distribution, then $p(x) \stackrel{d}{=} p(g_{\theta^*}(z))$, $\mathbb{E}_{p(z)}[f_\phi(g_{\theta^*}(z))] = \mathbb{E}_{p(x)}[f_\phi(x)]$ and:

$$\mathbb{E}[f_\phi(g_\theta(z))] - \mathbb{E}[f_\phi(x)] \approx \mathbb{E}[\nabla_\theta f_\phi(g_{\theta^*}(z))^\top (\theta - \theta^*)]$$

While MoLM optimizes a directional Jacobian of many moments, adversarial training optimizes a directional derivative for a single moment.

Similarly, one can think of discriminator penalties – such as gradient penalties in Wasserstein GANs (Gulrajani et al., 2017), the DRAGAN penalty (Kodali et al., 2017), and average second moment penalty Fisher GANs (Mroueh and Sercu, 2017) – as terms to control $\Sigma$. Of the three, Fisher GAN most directly controls the second moment by placing a penalty on $f_\phi^2$, but with respect to a mixture distribution of data and samples. That penalties such as the gradient penalty improve other adversarial objectives suggests a deeper connection to controlling $\Sigma$.

Also implied by adversarial training is that it matches at most one moment per generator step; due to space constraints, we defer discussion of asymptotics to Appendix B.3.

It is perhaps the connection to statistical efficiency, rather than the choice of a particular distance, that explains the success of adversarial training. It may also explain why direct minimization of Wasserstein distance in the primal yields much poorer samples than minimization in the dual. Developing this hypothesis may better explain why adversarial training works so well.

### 3.2.1. MOMENT MATCHING IN ADVERSARIAL NETWORKS

Moment matching has also found its way into adversarial training. Salimans et al. (2016) introduced feature matching of activations to stabilize GAN training. Mroueh et al. (2017) match mean and covariance embeddings of a neural network. Mroueh and Sercu (2017) also matches mean embeddings, but constrains the singular value of the covariance matrix of a mixture distribution between data and samples to be less than one. Li et al. (2017) attempts to learn invertible and adversarial feature mappings such that one can use an RBF kernel for MMD. Salimans et al. (2018) and Bellemare et al. (2017) also learn adversarial feature mappings in combination with Wasserstein and Energy distance, respectively. Most of these proposals, however, introduce too few moments to train a generator, and require frequent updates of features. The possible exception is MMD GAN, which tries to learn an invertible feature map that can be used with a kernel distance. Invertiblility is only enforced through regularization, and the feature mapping is likely not invertible in practice.

Table 1. Scores for different metrics for generators trained with random moments/MoLM. For Inception, higher scores are better; for FID, lower scores are better; and for MS-SSIM, scores closer to .379 are better.

| Metric/Dataset | CelebA | CIFAR-10 |
|---|---|---|
| Inception Score | - | 2.10/**6.99** |
| FID | - | 160.3/**33.8** |
| MS-SSIM | .444/**.378** | - |

## 4. Experimental Results

We evaluate our method on four datasets: Color MNIST (Metz et al., 2016), CelebA (Liu et al., 2015), CIFAR-10 (Krizhevsky, 2009), and the daisy portion of ImageNet (Russakovsky et al., 2015). We complement the visual inspection of samples with numerical measures to compare this method to existing work. For CelebA, we use Multi-Scale Structural Similarity (MS-SSIM) (Wang et al., 2003) to show sample similarity within a single class. Higher scores typically indicate mode collapse, while lower indicate higher diversity and better performance. Numbers lower than the test set may imply underfitting. For CIFAR-10, we include the standard Inception Score (IS) (Salimans et al., 2016) and Fréchet Inception Distance (FID) (Heusel et al., 2017).

We aim to answer three questions: 1) does learning moments improve sample quality, 2) what is the effect of including gradient and hidden unit features, and 3) how does sample quality compare to GAN alternatives?

We use convolutional architectures for both our generator and moment networks. To directly compare this algorithm to other methods, unless otherwise noted, we use a DCGAN generator. Direct comparisons using the same moment architecture as a discriminator makes less sense, however, since the set of moments used for adversarial learning come from an output of a network while our method uses low- and high-level information. Thus, we modify the architecture from a standard discriminator, though we only add size-preserving convolution before each stride-two layer of a DCGAN. For details of the specific architectures and hyperparameters used in all our experiments, please see Appendix A in the supplementary material. The models considered here are all unconditional.

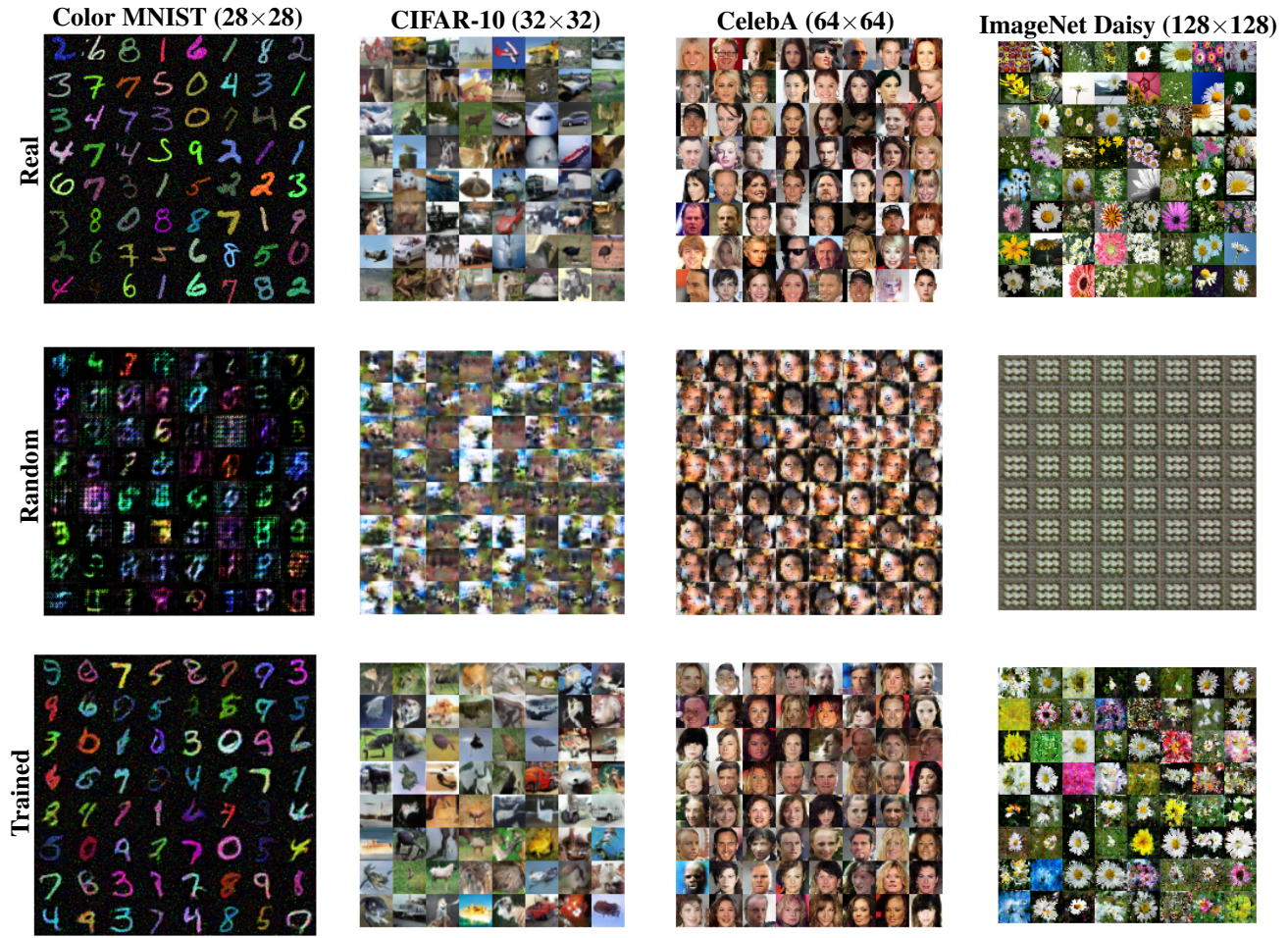

*Figure 2.* Top row are images from the the dataset (from left to right: Color MNIST, CIFAR-10, CelebA, and the Daisy portion of ImageNet at $128 \times 128$ resolution). The middle are samples trained with random moments. The bottom are sampled trained with MoLM.

To answer the first question – does learning moments improve sample quality –, we refer to Figure 2, which highlights the importance of learning moments. While random moments allow the generator to learn some structure of digits on Color MNIST, those moments are not sufficiently good to learn implicit generative models on other datasets. The sample quality is much higher for learned moments on all datasets. We also provide quantitative evidence for CIFAR-10 and CelebA: Table 1 shows a better Inception Score and Fréchet Inception Distance for learned moments compared to random moments on CIFAR-10, and a better MS-SSIM score on CelebA.

For the second question – what is the effect of including gradient and hidden unit features – we refer to Figure 4. For this experiment, we again focused on CIFAR-10 due to more robust metrics compared to other datasets. We tried four types of Adam hyperparameters, and two architectures. Across the board, we found that merely using activations did not work, which is not surprising as activations constitute roughly one-tenth the number of moments needed to train the generator.[3] Using only gradient features allows the model to learn more realistic samples. Using both, however, substantially improve IS and FID.

Finally, we compare Method of Learned Moments to Generative Adversarial Networks, and find MoLM performs as well as, if not better than, its GAN counterparts. On CelebA, shown in Table 2, the Multi-Scale Structural Similarity is as good as, or better than GAN alternatives. Admittedly, this metric is flawed as it only measures sample diversity and not quality of samples. At worst, however, the sample diversity of MoLM is comparable to GANs and the test set.

We find similar results on CIFAR-10. We try two convolutional architectures: the DCGAN, and one – denoted "Conv." – recently introduced in Miyato et al. (2018). As shown in Table 3, MoLM significantly outperforms gradient penalty and spectrally-normalized GANs on both Inception Score and Fréchet Inception Distance using the Conv. architecture. It also outperforms MMD alternatives using the DCGAN architecture, though the comparison is not entirely fair as the kernel sizes are different. Moreover, these latter results are fairly robust to increasing size of the moment network, shown in Figure 3. Inception Score and Fréchet Inception Distance also does not collapse over time.

---

[3]Please refer to Figure 5 in the Supplementary Material for samples for different types of moments.

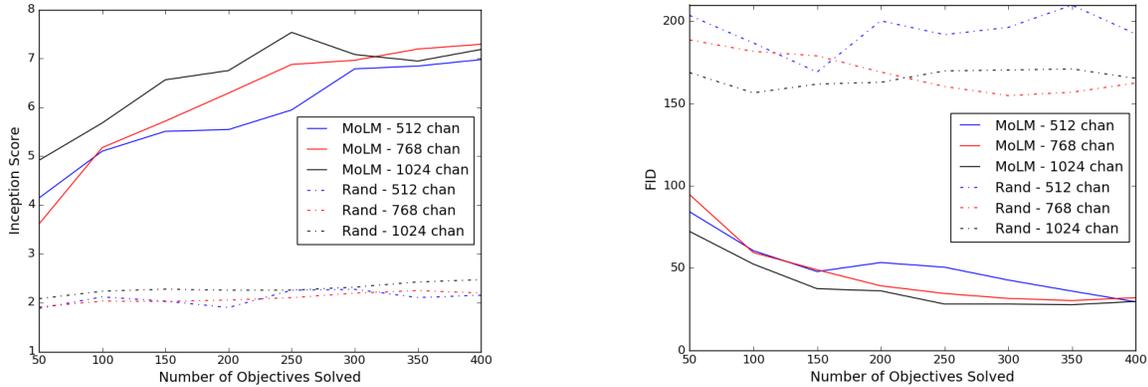

*Figure 3.* Left pane is Inception Score vs. and right is Fréchet Inception Distance vs. number of objectives solved for different size moment networks. For random moments, there is only a single objective but uses the same number of generator steps.

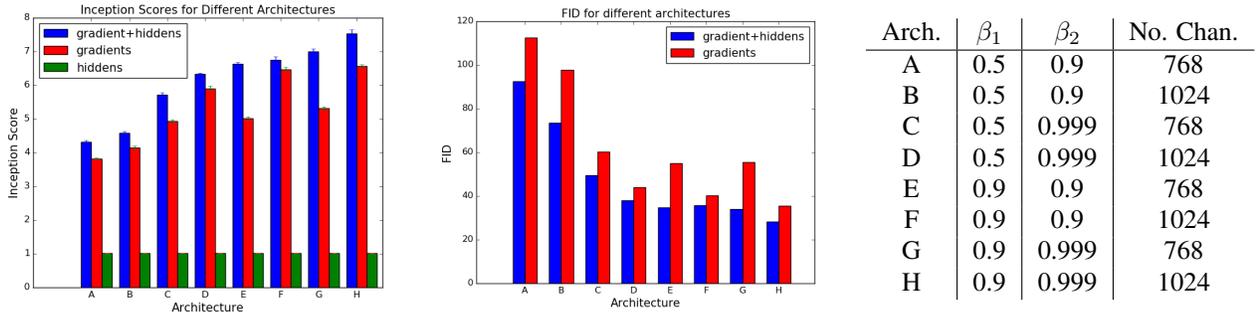

| Arch. | $\beta_1$ | $\beta_2$ | No. Chan. |
|---|---|---|---|
| A | 0.5 | 0.9 | 768 |
| B | 0.5 | 0.9 | 1024 |
| C | 0.5 | 0.999 | 768 |
| D | 0.5 | 0.999 | 1024 |
| E | 0.9 | 0.9 | 768 |
| F | 0.9 | 0.9 | 1024 |
| G | 0.9 | 0.999 | 768 |
| H | 0.9 | 0.999 | 1024 |

*Figure 4.* Inception Score and FID for different moment architectures and learning rates (right third). FID scores for just hiddens were not included because covariance matrix of the inception pool3 layer encountered rank deficiency.

*Table 2.* Multi-Scale Structural Similarity Results on CelebA for different methods.

| Method | GAN | GAN-GP | DRAGAN | WGAN-GP | MoLM (ours) | Test Set |
|---|---|---|---|---|---|---|
| MS-SSIM | .381 | .387 | .383 | .378 | .378 | .379 |

To illustrate that the method can scale to higher-resolution images, we generate examples from the daisy portion of ImageNet at 128×128 resolution. GANs currently perform conditional image generation on the full dataset; unfortunately, no such conditional version of MoLM currently exists. This preliminary result, however, demonstrates the promise of the algorithm.

## 5. Discussion

We introduce a method of moments algorithm for training large-scale implicit generative models. We highlight the importance of learning moments and create a stable learning algorithm that performs better than adversarial alternatives. The current algorithm, however, leaves some room for improvement. For example, the moment architectures, slightly modified from discriminator architectures used in adversarial learning, are likely suboptimal. Moreover, the current learning of moments relies on a binary classification heuristic that can almost certainly be improved.

*Table 3.* Comparison of Inception Scores (IS) and FID using 5,000 generated images(-5K) and 50,000 generated images (-50K) for different convolutional architectures. ⋆ from (Miyato et al., 2018). ∘ from (Li et al., 2017). Coulomb GAN † from (Unterthiner et al., 2017). Different methods are grouped by generator architecture.

| Arch. | Method | IS | FID-5K/50K |
|---|---|---|---|
| 4×4 Conv. | GAN-GP ⋆ | $6.93 \pm .11$ | 37.7/- |
| | WGAN-GP ⋆ | $6.68 \pm .06$ | 40.2/- |
| | SN-GAN ⋆ | $7.58 \pm .12$ | 25.5/- |
| | **MoLM-1024** | $\mathbf{7.55 \pm .08}$ | **25.0/20.3** |
| | **MoLM-1536** | $\mathbf{7.90 \pm .10}$ | **23.3/18.9** |
| 5×5 DCGAN | MMD-RBF ∘ | $3.47 \pm .03$ | -/- |
| | MMD-GAN ∘ | $6.17 \pm .07$ | -/- |
| | Coul. GAN † | - | -/27.3 |
| 4×4 DCGAN | **MoLM-768** | $\mathbf{7.56 \pm .05}$ | **31.4/27.3** |

Finally, a connection between adversarial learning and statistical efficiency seems to exist, and exploring this relationship may help us better understand the quiddity of GANs.

## Acknowledgments

We were incredibly lucky to have meaningful input that helped shape this work. We would like to thank David Warde-Farley, Aaron van den Oord, Razvan Pascanu, and the anonymous reviewers for trenchant insights, inspiration, and grammar corrections.

# Supplementary Material for Learning Implicit Generative Models with the Method of Learned Moments

## A. Experimental Details

### A.1. Experimental Setup

As mentioned in the main text, unless otherwise noted, generators use the standard DCGAN architecture with 4×4 kernels. The structure of the generator architectures for different datasets are described in Table 4.

The moment network for Color MNIST mirror the standard DCGAN discriminator architecture with one modification: after the last convolutional layer, we replace linear layer of size [4×4×C, 1] with two linear layers of sizes [4×4×C, noise dimension] and [noise dimension, 1], respectively, to ensure that there are at least as many moment network parameters as generator parameters. Furthermore, the generator is trained only with moments from gradient features, as activation features did not improve sample quality. This allowed the use of the Hessian-vector products to more quickly train the generator. Non-linearities between all layers are leaky ReLUs with leaky parameter 0.2.

For CIFAR-10, CelebA, and the daisy portion of ImageNet, we found some improvement using a larger moment network. Again, the moment network mirrors the DCGAN discriminator architecture, but with two changes: prior to each stride-2 convolutional layer we insert a stride-1 layer, and we decrease the kernel size to 3×3. Non-linearities between all layers are leaky ReLUs with leaky parameter 0.2. None of the moment networks use batch normalization. For experiments that used gradient and hidden unit features, hidden units were scaled by a constant factor (known as activation weights in Table 10) since the hidden units had a larger dynamic range than gradient features.

Table 10 shows the hyperparameters used for all experiments with a few exceptions. One is the the stability of MoLM training, which increases the number of objectives from 250 to 400. The second is the comparison of gradient features, activations, and both gradient features and activations, as we vary the size of the moment networks and vary the Adam optimizer's $\beta$ parameter in that experiment. The last is the comparison with GAN alternatives on CIFAR-10, and the differences are described in the last paragraph of Appendix A.1.

For comparisons, we use two standard, but somewhat flawed metrics: Inception Score (IS), and Fréchet Inception Distance (FID). For IS, we use the standard protocol and calculate scores using 10 batches of 5,000 images (for a total of 50,000) images. For FID, we report distances using 5,000 and 50,000 generated images for comparison with adversarial methods. For all CIFAR-10 experiments in the main text, we use ImageNet-trained networks, as this is the standard network for comparison. As noted by (Rosca et al., 2017; Barratt and Sharma, 2018), however, Inception Scores and Fréchet Inception Distances based on ImageNet-trained networks can lead to misleading results. Therefore, we also include Inception Scores on CIFAR-trained networks[4] in Table 8 for comparison with future work. N.B. we do not include FID results on CIFAR-trained networks, since FID for baseline and proposed methods are extremely low (less than 2.0). We surmise that this is the result of the embedding layer of the CIFAR-trained network being far lower-dimensional than that of the ImageNet-trained one.

On CelebA and CIFAR-10, we tried four GAN variants: GAN (Goodfellow et al., 2014) with and without a gradient penalty (Gulrajani et al., 2017), Wasserstein GAN with a gradient penalty (Gulrajani et al., 2017), and DRAGAN (Kodali et al., 2017) with nonsaturating loss. The same generator architecture was used for the GAN variants as MoLM. The results reported on DRAGAN, GAN-GP, and WGAN-GP were the best obtained in a hyperparameter sweep over discriminator learning rates in 0.0001, 0.0002, 0.0003 and generator learning rates in the same interval. Whenever applicable, the gradient penalty coefficient used was 10. The models were trained using the AdamOptimizer with $\beta_1$=0.5 and $\beta_2$=0.9. DRAGAN, GAN, and GAN-GP performed one discriminator update per generator update, while WGAN-GP performed 5 discriminator updates for generator updates, for a total of 200,000 generator updates.

On CIFAR-10, we found that our GAN variants had Inception Scores up to 0.2 worse than comparable published results. For completeness, we include these results in Table 7. We did not believe the this would be a reliable indicator of relative performance between adversarial methods and the proposed one. For a more sound comparison, we use GAN-GP and WGAN-GP results from Miyato et al. (2018), as those results are the best we found. It uses a different convolutional generator architecture (its specification can be found in Table 12), which provides the extra benefit of showing that MoLM can train more than just DCGAN generators. We also believe that those results are among the best

---

|  | Color MNIST | CIFAR-10 | CelebA | ImageNet Daisy |
|---|---|---|---|---|
| Noise dimension | 128 | 128 | 256 | 256 |
| Projection layer size | 4×4×256 | 4×4×512 | 4×4×512 | 4×4×512 |
| Conv. transpose layer 1 output size | 8×8×128 | 8×8×256 | 8×8×256 | 8×8×256 |
| Conv. transpose layer 2 output size | 16×16×64 | 16×16×128 | 16×16×128 | 16×16×128 |
| Conv. transpose layer 3 output size | N/A | N/A | 32×32×64 | 32×32×64 |
| Conv. transpose layer 4 output size | N/A | N/A | N/A | 64×64×32 |
| Output layer size | 32×32×3 | 32×32×3 | 64×64×3 | 128×128×3 |
| Output nonlinearity | sigmoid | tanh | tanh | tanh |
| Hidden nonlinearity | ReLU | ReLU | ReLU | ReLU |
| Kernel size | 5×5 | 4×4 | 4×4 | 4×4 |
| Batch norm | Yes | Yes | Yes | Yes |
| Number of parameters | 1,557,571 | 3,685,123 | 4,861,827 | 4,893,123 |

*Table 4.* Generator architectures across different datasets.

|  | MoLM-512 | MoLM-768 | MoLM-1024 | MoLM-1536 |
|---|---|---|---|---|
| Size-Preserving Layer 1 | 3×3×3×128 | 3×3×3×192 | 3×3×3×256 | 3×3×3×384 |
| Stride-2 Layer 1 | 3×3×128×128 | 3×3×192×192 | 3×3×256×256 | 3×3×384×384 |
| Size-Preserving Layer 2 | 3×3×128×256 | 3×3×192×384 | 3×3×256×512 | 3×3×384×768 |
| Stride-2 Layer 2 | 3×3×256×256 | 3×3×384×384 | 3×3×512×512 | 3×3×768×768 |
| Size-Preserving Layer 3 | 3×3×256×512 | 3×3×384×768 | 3×3×512×1024 | 3×3×768×1536 |
| Stride-2 Layer 3 | 3×3×512×512 | 3×3×768×768 | 3×3×1024×1024 | 3×3×1536×1536 |
| Linear Layer | 8,192 ×1 | 12,288×1 | 16,384×1 | 24,576×1 |
| Batch norm | No | No | No | No |
| Hidden nonlinearity | LReLU | LReLU | LReLU | LReLU |
| Number of Activations | 285,600 | 420,864 | 560,128 | 838,656 |
| Number of Parameters | 4,584,577 | 10,305,217 | 18,311,425 | 41,180,545 |
| Number of Total Moments | 4,866,177 | 10,726,081 | 18,871,553 | 42,019,201 |

*Table 5.* Moment Network Architectures for CIFAR-10

for GAN-GP and WGAN-GP for any generator architecture. We also compare to the spectrally-normalized GANs (SN-GAN) in that work. For the DCGAN generator, we compare against MMD-GAN and MMD-RBF as those can be considered moment-based methods. Results were taken from Li et al. (2017). Finally, we include published results for Coulomb GAN (Unterthiner et al., 2017).

### A.2. Large Generator Training on CelebA

The experiments in the main text only train generators with up to 5 million parameters. To show the method can scale to a larger number of generator parameters, we doubled the number of channels and increased the kernel size to 5×5. The number of parameters is now 20 million, and Table 11 details the architecture. The moment network mirrors a DCGAN discriminator with 1,024 channels, and adds an extra linear layer to ensure the number of moments is greater than the number of generator parameters. No hidden unit features were used in order to speed up training using the Hessian-vector product trick. Figure 6 shows the result of the experiment: while the generator surprisingly learns some structure of faces using random moments, the generator learns a higher-quality sampler of faces with MoLM.

## B. Consistency and Asymptotic Normality of Moment Estimators

In this section, we review the consistency and asymptotic normality conditions for moment estimators. Many of these conditions are now standard within a body of work in econometrics known as "Generalized Method of Moments."

### B.1. Consistency and Asymptotic Normality Conditions Squared Error Objective

The consistency and asymptotic normality conditions for the Equation 1 (reproduced below) are taken from (Hall, 2005).

$$\mathcal{L}^G(\theta) = m_N(x_{1,\ldots,N}, \theta)^\top W_N m_N(x_{1,\ldots,N}, \theta)$$

We remove the dependence on $\Phi$ because it is static. Note that below:

$$m(x, \theta) := m_1(x_1, \theta) = \Phi(x) - \mathbb{E}_{p(z)}[\Phi(g_\theta(z))]$$

Consistency conditions are:

|  | CelebA | ImageNet Daisy |
|---|---|---|
| Size-Preserving Layer 1 | 3×3×3×96 | 3×3×3×48 |
| Stride-2 Layer 1 | 3×3×96×96 | 3×3×48×48 |
| Size-Preserving Layer 2 | 3×3×96×192 | 3×3×48×96 |
| Stride-2 Layer 2 | 3×3×192×192 | 3×3×96×96 |
| Size-Preserving Layer 3 | 3×3×192×384 | 3×3×96×192 |
| Stride-2 Layer 1 | 3×3×384×384 | 3×3×192×192 |
| Size-Preserving Layer 4 | 3×3×384×768 | 3×3×192×384 |
| Stride-2 Layer 2 | 3×3×768×768 | 3×3×384×384 |
| Size-Preserving Layer 5 | N/A | 3×3×384×768 |
| Stride-2 Layer 5 | N/A | 3×3×768×768 |
| Linear Layer | 12,288×1 | 12,288×1 |
| Batch norm | No | No |
| Hidden nonlinearity | LReLU | LReLU |
| Number of Activations | 921,600 | 1,941,504 |
| Number of Parameters | 10,551,649 | 10,612,657 |
| Number of Total Moments | 11,473,249 | 12,554,161 |

*Table 6.* Moment Network Architectures for CelebA and ImageNet Daisy

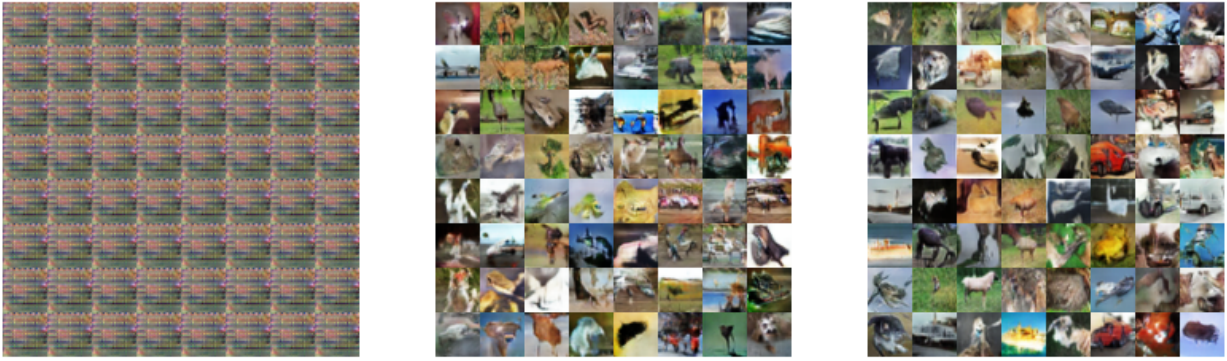

*Figure 5.* Samples for only activation features, gradient features, and gradient+activation features. Architecture and hyperparameters are using the default generator and MoLM-1024 moment network.

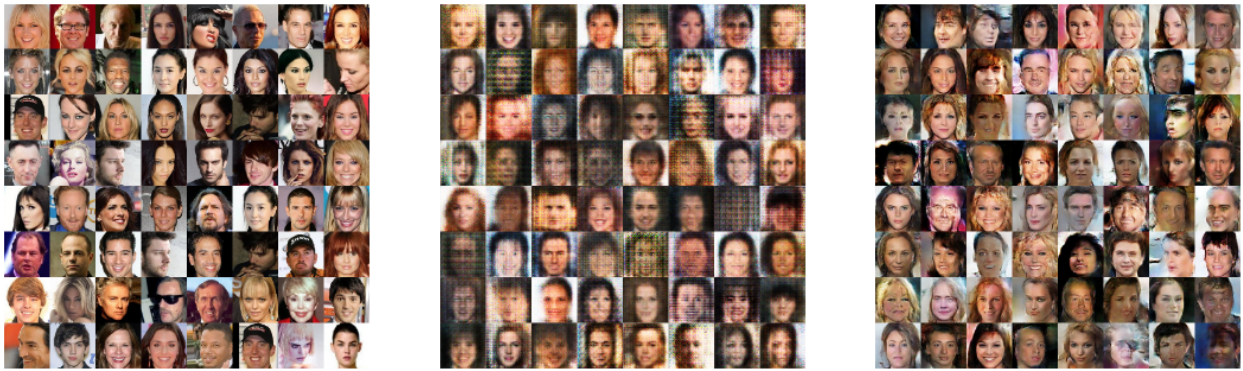

*Figure 6.* CelebA samples for large generator training. From left to right: 1) data, 2) examples from the generator trained with random moment network weights, 3) examples from the generator trained with MoLM.

- The $(d \times 1)$ random vectors $\{x_i; \ i = 1, \dots \}$ form a strictly stationary process with sample space $\mathbf{X} \subset \mathbb{R}^d$.
- The function $m : \mathbf{X} \times \Theta \to \mathbb{R}^k$, where $k < \infty$, satisfies: (i) it is continuous on $\Theta$ for each $x_i \in \mathbf{X}$; (ii)

$\mathbb{E}_{p(x)}[m(x, \theta)]$ exists and is finite for every $\theta \in \Theta$; (iii) $\mathbb{E}_{p(x)}[m(x, \theta)]$ is continuous on $\Theta$.
- The random vector $X$ and the parameter vector $\theta^*$ satisfy the population moment condition: $\mathbb{E}_{p(x)}[m(x, \theta^*)] = 0$

*Table 7.* Inception Score for baseline methods and MoLM on CIFAR-10.

| Method | Inception Score |
|---|---|
| GAN | 6.75 |
| GAN-GP | 6.88 |
| DRAGAN | 6.89 |
| WGAN-GP | 6.48 |
| **MoLM-768** | **7.56** |

*Table 8.* Inception Scores using a CIFAR-trained network for MoLM variants.

| Architecture | Method | Inception Score |
|---|---|---|
| DCGAN | GAN-GP | 6.41 |
| DCGAN | WGAN-GP | 6.34 |
| DCGAN | DRAGAN | 6.35 |
| **DCGAN** | **MoLM-768** | **6.55** |
| **Conv.** | **MoLM-1024** | **6.87** |
| **Conv.** | **MoLM-1536** | **7.13** |

and $\mathbb{E}_{p(x)}[m(x, \hat{\theta})] \neq 0 \quad \forall \hat{\theta} \neq \theta^*$.

- $W_N$ is a PSD matrix which converges in probability to the PD matrix of constants $W$.
- The random process $\{X_i, -\infty < i < \infty\}$ is ergodic.
- $\Theta$ is a compact set.
- $\mathbb{E}_{p(x)}[\sup_{\theta \in \Theta} \|m(X, \theta)\|] < \infty$

The third condition is known as global identifiability, and is typically difficult to verify. A heuristic that seems to work well in practice is to assume that the number of moments is greater than the number of model parameters, and that the Jacobian of moments with respect to the model parameters is full-rank.

If in addition the following conditions are true:

(I) (i) The derivative matrix $\nabla_\theta m(x_i, \theta)$ exists and is continuous on $\Theta$ for each $x_i \in X$; (ii) $\theta^*$ is an interior point of $\Theta$; (iii) $\mathbb{E}_{p(x)}[\nabla_\theta m(x, \theta^*)]$ exists and is finite.

(II) $\mathbb{E}_{p(x)}[m(x, \theta)m(x, \theta)^\mathsf{T}]$ exists and is finite, and $\lim_{T \to \infty} cov(T^{1/2} \sum_{i=1}^N \frac{m(x_i, \theta^*)}{N}) = \Sigma$ exists and is a finite valued positive definite matrix.

(III) $\mathbb{E}_{p(x)}[\nabla_\theta m(x, \theta)]$ is continuous on some neighborhood $\mathcal{N}_\epsilon$ of $\theta^*$.

(IV) $\sup_{\theta \in \mathcal{N}_\epsilon} \rho \xrightarrow{p} 0$ as $T \to \infty$.
$\rho = tr(\|\frac{1}{T} \sum_{i=1}^T \nabla_\theta m(x_i, \theta) - \mathbb{E}_{p(x)}[\nabla_\theta m(x, \theta)]\|^2)^{1/2}$

Then the estimator is asymptotically normal with variance given in Theorem 2.

## B.2. Consistency and Asymptotic Normality Conditions for Simulated Method of Moments

Duffie and Singleton (1993) proved consistency and asymp-totic normality for the more general case of Markov generators. In the i.i.d. scenario, some of the conditions are trivial. We modify the conditions for the i.i.d. case, but please refer to the original paper for more general conditions.

Consistency conditions are:

- For each $\theta \in \Theta, \{\|\Phi(g_\theta(z_i))\|_{2+\delta}, i = 1, 2, \dots\}$ is bounded for some $\delta > 0$. The family $\{\Phi(g_\theta(z_i))\}$ is Lipschitz, uniformly in probability.
- $\Sigma$ is nonsingular.
- Define $\mathcal{L}^G(\theta) = \hat{m}_N(x_{1,\dots,N}, \Phi, \theta)^\mathsf{T} \hat{m}_N(x_{1,\dots,N}, \Phi, \theta)$. Then $\mathcal{L}^G(\theta^*) < \mathcal{L}^G(\theta)$ for all $\theta \neq \theta^*$.

Asymptotic normality additionally requires:

- (i) $\theta^*$ and estimators $\{\hat{\theta}_N\}$ are interior to $\Theta$. (ii) $\Phi(g_\theta(z_i))$ is continuously differentiable with respect to $\theta$ for all $i$. (iii) $\mathbb{E}_{p(z)}[\nabla_\theta \Phi(g_{\theta^*}(z))]$ exists, is finite, and has full rank.
- The family $\{\nabla_\theta \Phi(g_\theta(z_i)), \theta \in \Theta, i = 1, 2, \dots\}$ is Lipschitz, uniformly in probability. For all $\theta \in \Theta, \mathbb{E}_{p(z)}[\|\nabla_\theta \Phi(g_\theta(z))\|] < \infty$, and $\theta \mapsto \mathbb{E}_{p(z)}[\nabla_\theta \Phi(g_\theta(z))]$ is continuous.

If the conditions are true, then the asymptotic variance is the one outlined in Theorem 3.

## B.3. Moment Matching with Alternative Distances

Adversarial training seems to be performing moment matching with access to a single moment per generator step. Can we say anything how this changes the asymptotics? Presently, no, but we can say something about the asymptotics of matching a finite number of moments with respect to another metric (in this case $\|l\|_\infty$), instead of squared error:

**Theorem 4.** *Under the Assumptions below, the estimator* $\hat{\theta}_N$ *converges in probability to* $\theta^*$. *Furthermore, we have:*

$$\sqrt{N}(\hat{\theta}_N - \theta^*) \to \arg\min_\zeta d(Y + G\zeta)$$

*where* $Y \sim \mathcal{N}(0, \Sigma)$ *and* $G := \mathbb{E}_{p(z)}[\nabla_\theta \Phi(g_{\theta^*}(z))]$

This result is proved in (Han and De Jong, 2004). Asymptotic normality requires conditions on the distance function $\delta(\cdot)$, conditions on the notion of a localized distance, and moment conditions. The conditions on the distance function are:

- $\delta(\cdot)$ is continuous
- $\delta(x) = 0$ iff $x = 0$
- $\delta(x) = \delta(-x)$

| | Color MNIST | CIFAR-10 | CelebA | ImageNet Daisy |
|---|---|---|---|---|
| Number of objectives $N_o$ | 150 | 250 | 250 | 250 |
| Number of moment training steps $N_m$ | 100 | 100 | 100 | 100 |
| Number of generating training steps $N_g$ | 2,000 | 2,000 | 2,000 | 2,000 |
| Learning rate $\alpha$ | 1E-4 | 1E-4 | 1E-4 | 1E-4 |
| Adam $\beta_1/\beta_2$ | 0.9/0.999 | 0.9/0.999 | 0.9/0.999 | 0.9/0.999 |
| Activation weights | 0.0 | 1E-4 | 1E-4 | 1E-4 |
| Norm penalty parameter $\lambda$ | 0.1 | 1.0 | 1.0 | 1.0 |
| Batch size | 1000 | 200 | 200 | 200 |

*Table 9.* Hyperparameters for different datasets for all experiments except those comparing to adversarial methods.

| | DCGAN | Conv |
|---|---|---|
| Number of objectives $N_o$ | 700 | 800 |
| Number of moment training steps $N_m$ | 50 | 50 |
| Number of generating training steps $N_g$ | 1,000 | 1,000 |
| Learning rate $\alpha$ | 1E-4 | 1E-4 |
| Adam $\beta_1/\beta_2$ | 0.9/0.999 | 0.9/0.999 |
| Activation weights | 1E-3 | 1E-3 |
| Norm penalty parameter $\lambda$ | 0.1 | 0.1 |
| Generator batch size | 200 | 200 |
| Moment batch size | 50 | 50 |

*Table 10.* Hyperparameters for different architectures for GAN comparison on CIFAR-10.

| | CIFAR-10 |
|---|---|
| Noise dimension | 128 |
| Projection layer size | 4×4×512 |
| Conv. transpose layer 1 output size | 8×8×256 |
| Conv. transpose layer 2 output size | 16×16×128 |
| Conv. transpose layer 3 output size | 32×32×64 |
| Stride-1 Conv. layer output size | 32×32×3 |
| Output nonlinearity | tanh |
| Conv. transpose layer kernel size | 4×4 |
| Stride-1 Conv. layer kernel size | 3×3 |
| Batch norm | Yes |
| Number of parameters | 3,811,907 |

*Table 12.* Generator architecture for GAN comparison on CIFAR-10.

| | CelebA |
|---|---|
| Noise dimension | 256 |
| Projection layer size | 4×4×1024 |
| Conv. transpose layer 1 output size | 8×8×512 |
| Conv. transpose layer 2 output size | 16×16×256 |
| Conv. transpose layer 3 output size | 32×32×128 |
| Output layer size | 64×64×3 |
| Output nonlinearity | tanh |
| Kernel size | 5×5 |
| Batch norm | Yes |
| Number of parameters | 20,615,427 |

*Table 11.* Generator architecture for large generator parameter experiment.

- $\delta$ satisfies the triangle inequality up to a finite constant locally (in a neighborhood of 0), i.e., there exists an $\epsilon > 0$ such that if $\|x_1\|_1 < \epsilon$ and $\|x_2\|_1 < \epsilon$ then $\delta(x_1 + x_2) \leq M[\delta(x_1) + \delta(x_2)]\ \forall x_1, x_2$, for some $M < \infty$.

The authors define a sequence of *localized distance* functions as

$$d_n(x) = \frac{\delta(n^{-1/2}x)}{\delta(n^{-1/2}1)}\ \ n = 1, 2, \ldots$$

Conditions on the localized distance are:

- There is a real function $\phi(\cdot)$ on $\mathbb{R}^q$ such that $\inf_n d_n(x) \geq \phi(x)$, and $\phi(x) \to \infty$ if $|x| \to \infty$.
- $d_n$ converges uniformly in every compact subset of $\mathbb{R}^q$ to a continuous function $d$.
- $d(z + Bt)$ achieves its minimum at a unique point of $t \in \mathbb{R}^p$ for each $z \in \mathbb{R}^q$ and for any $q \times p$ matrix $B$ with full column rank.

Conditions on the moments (again removing dependence on $\Phi$) are:

- $\Theta$ is a compact set.

- $\hat{m}_N(x_{1,\ldots,N}, \theta) = \frac{1}{N} \sum_{i=1}^{N} m(x_i, \theta)$ converges in probability to a nonrandom function $\mu(\theta)$ uniformly on $\Theta$.
- $\mu(\theta) = 0$ iff $\theta = \theta^*$ where $\theta^*$ is an interior point of $\Theta$.
- $\hat{G}_N(\theta) = \frac{1}{N} \sum_{i=1}^{N} \nabla_\theta m(x_i, \theta)$ exists and converges in probability to a nonrandom function $G(\theta)$ uniformly in a neighborhood of $\theta^*$ and $G(\theta^*)$ has full column rank.
- There exists $\hat{\theta}$ in between $\theta$ and $\theta^*$ such that .

$$\hat{m}_N(x_{1,\ldots,N}, \theta) = \hat{m}_N(x_{1,\ldots,N}, \theta^*) + \hat{G}_N(\theta)(\theta - \theta^*)$$

for $\theta$ in a neighborhood of $\theta^*$.

- $\sqrt{N} \hat{m}_N(x_{1,\ldots,N}, \theta) \xrightarrow{d} \mathcal{N}(0, \Sigma)$

# C. Proofs

## C.1. Proof of Proposition 1

The proof of the following statement is sufficiently simple that there is likely an earlier proof. Unfortunately, we could not find a reference, so we are likely re-proving this statement.

**Proposition.** *Suppose the kernel function* $K(x, y) = K(x - y)$ *is real, shift-invariant, Bochner integrable, and without loss of generality* $K(0)=1$. *Then:*

$$\mathbb{E}_{p(x,x')}[K(x, x')] - 2\mathbb{E}_{p(x,y)}[K(x, y)] + \mathbb{E}_{p(y,y')}[K(y, y')]$$
$$= \mathbb{E}_{p(w)}[(\mathbb{E}_{p(x)}[\cos(\omega^\mathsf{T} x)] - \mathbb{E}_{p(y)}[\cos(\omega^\mathsf{T} y)])^2]$$
$$+ \mathbb{E}_{p(w)}[(\mathbb{E}_{p(x)}[\sin(\omega^\mathsf{T} x)] - \mathbb{E}_{p(y)}[\sin(\omega^\mathsf{T} y)])^2]$$

*where* $p(\omega)$ *is a probability measure specified by the kernel.*

*Proof.* From Bochner's Theorem for real kernels (Zhao and Meng, 2015):

$$K(x - y) = \mathbb{E}_{p(\omega)}[K(0) \cos(\omega^\mathsf{T}(x - y))]$$

When $K(0) = 1$, $p(\omega)$ is a probability measure. Without loss of generality let $K(0) = 1$. Since the kernel is integrable we can interchange expectations.

$$\mathbb{E}[K(x, y)] = \mathbb{E}_{p(x,y)}[\mathbb{E}_{p(\omega)}[\cos(\omega^\mathsf{T}(x - y))]]$$
$$= \mathbb{E}_{p(\omega)}[\mathbb{E}_{p(x,y)}[\cos(\omega^\mathsf{T}(x - y))]]$$

Then:

$$\mathbb{E}_{p(x,y)}[\cos(\omega^\mathsf{T}(x - y))] = \mathbb{E}_{p(x,y)}[\cos(\omega^\mathsf{T} x) \cos(\omega^\mathsf{T} y)]$$
$$+ \mathbb{E}_{p(x,y)}[\sin(\omega^\mathsf{T} x) \sin(\omega^\mathsf{T} y)]$$
$$= \mathbb{E}_{p(x)}[\cos(\omega^\mathsf{T} x)]\mathbb{E}_{(y)}[\cos(\omega^\mathsf{T} y)]$$
$$+ \mathbb{E}_{p(x)}[\sin(\omega^\mathsf{T} x)]\mathbb{E}_{p(y)}[\sin(\omega^\mathsf{T} y)]$$

Addition of $\mathbb{E}_{p(x,x')}[K(x, x')] - 2\mathbb{E}_{p(x,y)}[K(x, y)] + \mathbb{E}_{p(y,y')}[K(y, y')]$ yields the result. $\square$

## C.2. Simplification of Coulomb GAN

We offer a simpler interpretation of optimality of the generator in Coulomb GAN (Unterthiner et al., 2017) using ideas from Maximum Mean Discrepancy. Suppose we are learning an implicit generative model using MMD:

$$\mathcal{L}(\theta) = \min_\theta \sup_{f \in \mathcal{F}} \mathbb{E}_{p(x)}[f(x)] - \mathbb{E}_{p(z)}[f(g_\theta(z))]$$

If we knew $f^*$, the function that maximizes the inner supremum, then we can simplify the loss to:

$$\mathcal{L}(\theta) = \min_\theta -\mathbb{E}_{p(z)}[f^*(g_\theta(z))] \tag{3}$$

If the function class $\mathcal{F}$ is the unit ball in a Reproducing Kernel Hilbert Space, then the witness function $f^*$, defined in Gretton et al. (2012), can be analytically calculated as:

$$f^*(t) \propto \mathbb{E}_{p(x)}[k(x, t)] - \mathbb{E}_{p(z)}[k(g_\theta(z), t)]$$

The empirical version of which is:

$$\hat{f}^*(t) \propto \frac{1}{m} \sum_i k(x_i, t) - \frac{1}{n} \sum_j k(g_\theta(z_j), t)$$

Plugging in this scaled witness function into the Monte Carlo estimate of Equation 3 gives us a biased estimate of the loss. $\mathcal{L}(\theta)$ is a distance if the kernel $k(x, y)$ is characteristic.

In Coulomb GAN (Unterthiner et al., 2017), the discriminator and generator steps are:

$$\mathcal{L}_D(D; G) = \frac{1}{2}\mathbb{E}_{p(t)}\left((D(t) - \hat{\Phi}(t))^2\right)$$
$$\mathcal{L}_G(D; G) = -\frac{1}{2}\mathbb{E}_{p(z)}(D(g_\theta(z)))$$

The authors define the empirical estimate of the potential function $\Phi$ (not to be confused with feature functions in the main text) as:

$$\hat{\Phi}(t) = \frac{1}{m} \sum_i k(x_i, t) - \frac{1}{n} \sum_j k(g_\theta(z_j), t)$$

and

$$p(t) = \frac{1}{2} \int \mathcal{N}(t; g_\theta(z), \epsilon I)p_z(z)dz$$
$$+ \frac{1}{2} \int \mathcal{N}(t; x, \epsilon I)p_x(x)dx$$

$\hat{\Phi}$ is merely the empirical estimate of the witness function, discriminator $D$ is a model of the empirical witness function, and the generator loss is that of Equation 3. The empirical estimate of $\mathcal{L}_G(D; G)$ is biased, though it's unknown how this affects training in practice. Note that dependence of $\theta$ on $f^*$ requires frequent retraining of $D$.

To demonstrate that the loss is a distance, it remains to show that the function class $\mathcal{F}$ is rich enough, or equivalently that the kernel function $k(x, y)$ is characteristic. Note that the proposed Plummer kernel:

$$k_p(a, b) = \frac{1}{(\sqrt{\|a - b\|^2 + \epsilon^2})^d}$$

is a rational quadratic kernel:

$$k_{rq}(a, b) = \sigma^2 \left( 1 + \frac{\|a - b\|^2}{2\alpha l^2} \right)^{-\alpha}$$

with $\alpha = \frac{d}{2}$, $\sigma = \epsilon^{-d/2}$ and $l = \frac{\epsilon}{\sqrt{d}}$. Since rational quadratic kernels are characteristic, so are Plummer kernels.